\documentclass[10pt,journal,compsoc]{IEEEtran}
\ifCLASSOPTIONcompsoc
  \usepackage[nocompress]{cite}
\else
  \usepackage{cite}
\fi
\ifCLASSINFOpdf
\else
\fi

\usepackage{amsmath}
\usepackage{hyperref}
\newtheorem{definition}{Definition}

\usepackage{color}
\usepackage[dvipsnames]{xcolor}
\usepackage{graphicx}
\usepackage{subfigure}
\usepackage{multirow}
\usepackage{amssymb}
\usepackage{array}
\usepackage{threeparttable}
\usepackage{colortbl}
\usepackage{cite}
\usepackage{times}
\usepackage{latexsym}
\usepackage{booktabs}
\usepackage[ruled,linesnumbered]{algorithm2e}
\definecolor{mygray}{gray}{0.85}
\newcommand{\tabincell}[2]{\begin{tabular}{@{}#1@{}}#2\end{tabular}}

\begin{document}

\title{Recognizing Conditional Causal Relationships about Emotions and Their Corresponding Conditions}

\author{Xinhong~Chen,
        Zongxi~Li,~\IEEEmembership{Member,~IEEE},
        Yaowei~Wang,
        Haoran~Xie,~\IEEEmembership{Senior Member,~IEEE}
        Jianping~Wang,~\IEEEmembership{Fellow,~IEEE},
        and~Qing~Li,~\IEEEmembership{Fellow,~IEEE}
\IEEEcompsocitemizethanks{
\IEEEcompsocthanksitem Xinhong Chen and Jianping Wang are with the Department of Computer Science, City University of Hong Kong, Hong Kong, China.
\IEEEcompsocthanksitem Zongxi Li is with the School of Science and Technology, Hong Kong Metropolitan University, Kowloon, Hong Kong, China.
\IEEEcompsocthanksitem Haoran Xie is with the Department of Computing and Decision Sciences, Lingnan University, Hong Kong.
\IEEEcompsocthanksitem Yaowei Wang and Qing Li are with the Department of Computing, Hong Kong Polytechnic University, Hong Kong, China.
}}

\markboth{Journal of \LaTeX\ Class Files,~Vol.~14, No.~8, August~2015}%
{Chen \MakeLowercase{\textit{et al.}}: Bare Demo of IEEEtran.cls for Computer Society Journals}

\IEEEtitleabstractindextext{%
\begin{abstract}
The study of causal relationships between emotions and causes in texts has recently received much attention. Most works focus on extracting causally related clauses from documents but neglect the fact that the extracted emotion and cause clauses can only be valid under some specific context. To highlight the context in such special causal relationships, we propose a new task to (1) determine whether an input pair of emotion and cause has a valid causal relationship under different contexts and (2) extract the specific context clauses that participate in the causal relationship. Since no existing dataset is available for the new task, we manually annotate each context clause's type on a benchmark dataset. Negative sampling is employed to balance the number of documents with and without causal relationships in the dataset. Based on the constructed dataset, we propose an end-to-end multi-task framework with two novel modules for the proposed tasks: a context masking module extracting the context clauses participating in the causal relationships and a prediction aggregation module fine-tuning the predictions according to the dependency between the extracted causal clauses and the specific context clauses. Results of extensive comparative experiments and ablation studies demonstrate the effectiveness of our proposed framework.

\end{abstract}

}

\maketitle

\IEEEdisplaynontitleabstractindextext
\IEEEpeerreviewmaketitle

\IEEEraisesectionheading{\section{Introduction}\label{sec:intro}}
\IEEEPARstart{R}{ecently}, the research on the causal relationships between human emotions and their corresponding causes has received much attention. Recognizing the causes of a specific emotion in a document is considered as more useful than only identifying the emotion, due to the great potential of helping people make reasonable decisions and avoid unnecessary loss \cite{gui2017,li2018,xia2019,ding:19}. 

There are currently two main tasks concerning the causal relationships between emotions and their causes, the Emotion Cause Extraction (ECE) task \cite{lee2010,chen2010,gui2016a,gui2016b} and the Emotion-Cause Pair Extraction (ECPE) task \cite{xia:19,chen-etal-2018-joint}. Specifically, the ECE task focuses on extracting the causes for a given emotion, while the ECPE task focuses on extracting emotions and the corresponding causes as pairs. 

\begin{table}
\centering 
\caption{An example document where the context clauses participate in the causal relationship}
\label{tab:exampledoc}
\renewcommand\arraystretch{1.4}
\resizebox{80mm}{20mm}{
\begin{tabular}{l}  
\toprule
Document content \\ 
\midrule
\tabincell{l}{\textcolor{OliveGreen}{(1) Wu was diagnosed with advanced liver cancer in early 2014}, \\ \textcolor{OliveGreen}{(2) since when he began to update his health condition in Microblog}. \\ \textcolor{red}{(3) If Wu didn't update his microblog for a long time}, \\ \textcolor{blue}{(4) people worried that he may have passed away}. \\ (5) By the end of 2013, \\ (6) Wu hadn't updated his microblog account for over two months. \\  $...$ \\ } \\
\bottomrule

$^*$context clauses: (1), (2), (5), (6); cause clause: (3); emotion clause: (4) \\
\end{tabular}
}
\end{table}

Despite gaining increasing attention from academia, both tasks only aim to extract the clauses containing causal relationships and have neglected that some context clauses may be indispensable for the extracted clauses to have a valid causal relationship. Let us consider an example as shown in Table \ref{tab:exampledoc}. In the example, the red clause is the cause clause, the blue clause is the emotion clause, and the rest are context clauses. The red cause clause and the blue emotion clause may not have a causal relationship if we ignore the green context clauses, since the reasons of not updating one's social media account can be more than the owner passing away, such as forgetting his/her passwords, using a new account, etc. Only when extra information in the green context clauses is available, can these two clauses have a valid causal relationship? Such a special case is referred to as a \textit{conditional causal relationship} in this paper, which will be formally defined in Section \ref{sec:problem}, and the ``condition'' in the above example is the two green context clauses.

From the above example, we can see that it is essential to take the context clauses into consideration when determining the causal relationships between the emotion and cause clauses. With these conditional causal relationships figured out, more complete and meaningful information about the causal relationships can be extracted. Specifically, we can learn that some types of events may evoke different types of emotions under different circumstances. For example, generally one shall feel happy when he/she can have a seafood dinner, but someone allergic to seafood will feel sad about having a seafood dinner since he/she cannot enjoy the meal. Such context-oriented information can be beneficial in many emotion-related applications, such as accurately predicting one’s emotions with context taken into consideration when a specific event occurs.

Although the conditional causal relationships described above are ubiquitous in textual materials, no existing work has paid attention to them. In this paper, we articulate the importance of context in causal relationship recognition and make the first step of studying such special causal relationships in text data. We propose a new task to determine whether or not the input pair of emotion and cause clauses has a causal relationship given a specific context, and extract the specific context clauses that participate in the causal relationship. 

As our task is new without any existing annotations available, we conduct manual annotation on the ECPE dataset constructed by \cite{xia:19}, in which the emotions and causes are already annotated for each document. Specifically, we recruit three human experts from the areas of sentiment analysis and emotion classification to annotate the documents with the labels of conditional causal relationships and their corresponding context clauses. When annotating the context clauses, instead of using a binary label, we define two types of context clauses to distinguish the different roles of context clauses, namely \textit{Irrelevant (IR)} and \textit{Emotion-Cause-Pair-Related (PR)}. Specifically, a \textit{IR} context clause is irrelevant to the causal relationship and a \textit{PR} context clause is related to both the annotated cause and emotion. According to whether an annotated emotion-cause pair (ECP) is conditional or not, the \textit{PR} type context clauses can have different roles and hence can be used in different tasks. More details can be found in Section \ref{sec:contextrole}.To sum up, during annotation, the three experts are required to: (1) judge whether the annotated causes and emotions in a document need some specific context clauses to have a valid causal relationship; and (2) go through all context clauses one by one to determine their specific types. After annotation, we follow the procedure of negative sampling \cite{mikolov2013} to construct our final dataset, where we define two types of negative samples and different sampling operations for documents with a conditional causal relationship and those with a non-conditional one.

To handle our new task, we propose an end-to-end multi-task framework, in which we design two novel and general modules. The first module is a context masking module, which learns a binary mask over all the context clauses and aims to extract the ones that participate in the causal relationship. The other one is a prediction aggregation module designed to deal with the recognition of both conditional and non-conditional causal relationships, which automatically fine-tunes the prediction results by aggregating the predictions with and without context information.

The contributions of this work can be summarized as follows.

(1) To address the issue that some context clauses may be indispensable for a causal relationship to be valid, we define a new task to determine whether or not an input pair of emotion and cause has a causal relationship under a given context, and extract the specific context clauses that participate in the causal relationships.

(2) Based on the ECPE dataset, we construct a dataset via manual annotation and negative sampling, which can not only be used for our proposed task but also be applicable to other important applications.

(3) We propose an end-to-end multi-task framework with two newly proposed modules to handle our proposed task, including a context masking module and a prediction aggregation module. Results of extensive experiment and ablation studies demonstrate the effectiveness and generality of our proposed modules and framework.

An early stage of this work has been published in \cite{chenemnlp}, and in this paper, we expand that work in terms of both technical and experimental contents. The main enhancements include: (1) we provide more fine-grained annotations concerning conditional causal relationships between emotions and causes, including the labels of the conditional causal relationships and also different types of context clauses; (2) we propose a context masking module to achieve the goal of finding the specific context clauses that participate in the conditional causal relationships; (3) we conduct more extensive experiments and ablation studies to demonstrate the effectiveness and generality of our proposed modules.

The rest of the paper is organized as follows. We first review recent works concerning causal relationship recognition and the role of context in the causal relationships in Section \ref{sec:relatedwork}. In Section \ref{sec:problem}, we define the conditional causal relationship in terms of the ``emotion-cause pair'' proposed in the original ECPE task \cite{xia:19} and formulate our new task accordingly. Then, we describe the specific procedures of constructing a dataset for our proposed task based on the benchmark ECPE dataset in Section \ref{sec:dataset}. Section \ref{sec:model} presents our multi-task framework with the two newly proposed modules for our task, and Section \ref{sec:experiment} reports the results of both comparative experiments and ablation studies. Finally, we conclude our work and discuss some future directions in Section \ref{sec:conclusion}.

\section{Related Works}
\label{sec:relatedwork}
In this section, we review some recent efforts on recognizing causal relationships in text data, and also some literature about the role of context in the causal relationships.

\subsection{Causal Relationship Recognition}
The tasks concerning causality extraction in texts can be mainly divided into two categories, causal phrase extraction, and causal clause extraction. The causal phrase extraction task aims to extract word phrases that have causal relationships in a sentence, while the causal clause extraction task mainly focuses on extracting multiple clauses with causal relationships from a document. 



Two representative tasks of causal clause extraction are the ECE task \cite{lee2010,chen2010,gui2016a,gui2016b} and the ECPE task \cite{xia:19,chen-etal-2018-joint,wei-etal-2020-effective,TANG2020329,song2020endtoend}. The emergence of these two tasks can be traced back to the significant meaning of studying human sentiments and emotions, which is beneficial to help machines understand human beings' current status for better decision-making and customized product design \cite{BASIRI2021279}. Different research directions concerning human emotion and sentiments have already been proposed and broadly studied, including text classification \cite{royaaai}, emotion classification \cite{roykbs,royec,tacec1}, sentiment analysis \cite{erikemotion,tacsa1}, sentiment dictionary construction \cite{senticnet}, etc. ECE and ECPE are two recently proposed tasks that aim to study the causal relationships between human emotions and their corresponding causes. Existing works of these two tasks are mostly based on deep learning classification-based models \cite{eriksurvey} to extract abstract features for each clause and aim to classify whether or not some clauses are causally related.

Although the context of the input document is always involved in providing more semantic information to enhance the clauses' embedding vectors \cite{KruengkraiTHKOT17,keyash2019,zhou2016,LI2019512}, none of the above works have paid attention to the possible effect of context on the causal relationship itself. Moreover, as our focus in this paper is on emotion-causal relationships, for some emotions (e.g., \textit{shame}, \textit{envy}, \textit{guilt}, etc.) to arise in the first place, a particular social setting may be necessary. Therefore, taking the social contexts into consideration may be an essential step to study whether a specific event can cause an emotion \cite{wendy2015,marsella2010computational,jurafsky200426}. 

In this work, we articulate the importance of context in causal relationship recognition, in view that some context clauses may be essential in order for a pair of emotions and cause to have a valid causal relationship.

\subsection{The Role of Context}
In traditional causal reasoning, the term ``context'' is mostly discussed in the task of \textit{causal effect estimation}, which is to estimate the influence of the cause variable on the effect variable \cite{guo2018survey}. In some cases, both of the cause and effect variables can be simultaneously affected by some other variables called \textit{confounders} \cite{guo2018survey}, which can be viewed as the context variables. These confounders should be discovered and ``removed'' by some specially designed algorithms in order to accurately estimate the causal effect or discover correct causal relationships \cite{multipleconfound,ruichu2,causalinference1}, such as propensity score method \cite{xingsamgu1993,peterc2011,Lunceford2004}, front-door criterion \cite{pearl1995}, instrumental variable estimator combined with structural causal models or potential outcome framework \cite{guo2018survey}, etc.


In contrast, our focus is on another role of context in the causal relationship, \textit{condition}. Unlike confounders described above and  \textit{mediator} defined in traditional causal reasoning \cite{guo2018survey} that aims to model the indirect causal effect between the cause and effect variables, in this paper, the condition context in texts refers to an essential set of context clauses for a pair of cause and emotion to have a causal relationship. Although mediator and condition seem to share some common properties, there is a fundamental difference between these two concepts. In general, a mediator can be affected by the cause variable and then pass the influence to the effect variable, while a condition may only influence the relationship itself instead of either the cause or the effect.

Currently, there is no existing work concerning causal relationships in text data discussing the condition context, although the conditional causal relationship can be observed in most textual materials in the real world. In this paper, we aim to take the first step to study such special causal relationships by defining a new task in the following section.

\section{Task Definition}
\label{sec:problem}
In this section, we first formally define the term ``conditional emotion-cause pair'' based on the notion of ``emotion-cause pair'' proposed in the original ECPE task \cite{xia:19}, and then formulate our proposed task accordingly. 

As defined by \cite{xia:19}, an emotion-cause pair (ECP) contains an emotion clause indicating an emotion (e.g., \textit{Happiness}) and a set of corresponding cause clauses. For an ECP in a document, except for the clauses contained in the ECP, the remaining clauses will be referred to as its \textit{context clauses}. With the above definition of ECP and context clauses, we define ``conditional ECP'' as follows.

\begin{definition}[\textbf{Conditional Emotion-Cause Pair}]
\label{def:cecp}
\textit{If an emotion-cause pair is considered to have a causal relationship only when some specific context clauses are given, it is called a conditional emotion-cause pair.}
\end{definition}


An example document with a conditional pair and its explanation can be found in Table \ref{tab:exampledoc} and Section \ref{sec:intro}. 
Definition \ref{def:cecp} indicates that a conditional pair should not be judged to have a causal relationship when the given context does not contain the information it needs. Also, according to whether or not a context clause is needed by the ECP, we can change the goal of finding specific participating context clauses into predicting a binary mask over all the context clauses, where $1$ indicates a context clause participating in the causal relationship and $0$ indicates that it is irrelevant to the causal relationship. Based on the above description, our task is formulated as follows.

\textbf{The proposed task}: Given a specific context $con_i$ and an emotion-cause pair $x_i=(c_i, e_i)$ containing a set of cause clauses $c_i$ and an emotion clause $e_i$, determine a binary label $y_i$ indicating whether or not the input pair $x_i$ has a causal relationship under the context $con_i$, and a binary mask $y^m_i$ over $con_i$ to indicate the context clauses that actually participate in the causal relationship.

As defined above, our proposed task is not to directly distinguish the conditional pairs from the non-conditional ones. The reason is that the recognition of a conditional pair is based on its different labels under different contexts instead of the text itself. Therefore, such a task formulation spares the models from worrying about how to transform the labels of causal relationships to the labels of conditional pairs and simplifies the process of training. 

\section{Dataset Construction}
\label{sec:dataset}
Since our proposed task is new for which no existing annotation is available, we construct our own dataset based on the ECPE dataset \cite{xia:19} through two steps: manual annotation and negative sampling. 

\subsection{Manual Annotation}
\label{sec:manualanno}
In the ECPE dataset, documents are mainly snippets of news articles or social media articles, which may contain multiple ECPs. As we aim to analyze whether each ECP is conditional or not, for each document with multiple ECPs, we duplicate it multiple times and only keep the annotation of one ECP for each duplicated document. By doing this, we assure that each document in the dataset will only contain one ECP. 

To handle our proposed task, we need the label of conditional pair for each document, and the labels of context clauses indicating whether or not they are essential to the causal relationship. Therefore, our manual annotation includes the following two steps: \textit{conditional ECP annotation} and \textit{context clause annotation}. 

We recruit three human experts who are experienced academic partners in the area of emotion classification and sentiment analysis to help us conduct the manual annotation. The three experts are required to (1) determine the labels of conditional ECP for each document to be described in Section \ref{sec:condlabel}, and (2) assign each context clause with a label type to be defined in Section \ref{sec:contextrole}. Note that the three annotators are required to coordinate their annotation standards before their own annotation to reduce any form of bias (e.g., confirmation bias) to the largest degree. After the annotation process, instances with inconsistent labels will be further discussed among the annotators. In what follows, we describe the two steps of manual annotation in detail.

\subsubsection{Conditional ECP Annotation}
\label{sec:condlabel}
To label an ECP as a conditional one, the cause events and the effect emotions should not be causally related when given an irrelevant context. For example, someone allergic to seafood may feel down since he/she cannot enjoy the seafood. Such context information outside of the cause and emotion clauses is what the three experts are required to find. 

In most documents with conditional ECPs, the context information needed by the ECP is contained in the context clauses of the document. However, it is still possible that the desired information is not contained in the current document, since these documents may originally be small paragraphs of a longer article and separated into multiple pieces in the original ECPE dataset. Taking such a possibility into consideration, we require the three experts to give a separate label to the documents in which the context information needed for the conditional ECP is missing.

To sum up, during the annotation process of conditional ECP, the three experts need to (1) give a binary label of conditional ECP $y_{ce}$ to each document in the original ECPE dataset, indicating whether the ECP is a conditional one; and then (2) if $y_{ce} = 1$, give another binary label $y_{cv}$ to indicate whether the needed context information for the conditional ECP is included in the documents.
Specifically, for the label of conditional ECP, $y_{ce} = 1$ indicates that the ECP in the document is conditional and $y_{ce} = 0$ indicates that it is not. As for another binary label, $y_{cv} = 1$ indicates the context information needed for the conditional ECP is included in the document and $y_{cv} = 0$ indicates that it is not. 

To validate the fidelity of the manual labels provided by the three experts, we calculate the average agreement rate of $y_{ce}$ and $y_{cv}$ among the annotators, which are \underline{\textbf{91.53\%}} and \underline{\textbf{95.25\%}}, respectively. Therefore, these labels are credible and can be used for our proposed task.

Since there are three experts and two binary labels, we can have a maximum of 6 labels for a document. To aggregate the labels from the three experts, for each document we first apply majority voting separately for the two binary labels. For example, given a document, if two experts agree on $y_{ce} = 1$, then the aggregated label of $y_{ce}$ for this document is 1. After we get two aggregated binary labels for each document, we combine them to get the final label $y_c$ as $y_c = y_{ce} + y_{ce} * y_{cv}$,
where $y_c = 0$ indicates the ECP of the document is non-conditional, $y_c = 1$ indicates the ECP of the document is conditional but the context information needed is missing, and $y_c = 2$ indicates the ECP is conditional and the context information it needs is included in the document. Accordingly, the label of causal relationship for each document $y_i$ is set to 1 when $y_c$ is 0 or 2, and when $y_c$ is 1, $y_i$ is 0.


\subsubsection{Context Clause Annotation}
\label{sec:contextrole}
After labeling an ECP as a conditional one, the next step is to find the corresponding context clauses related to the ECP. 

To achieve such a goal, we define two types of context clauses and the three experts need to assign each context clause to one of the two types. Specifically, we define \textbf{Irrelevant (IR)} type of context clause as those irrelevant to the causal relationship contained in the document and \textbf{ECP-related (PR)} type of context clause as those directly related to the annotated ECP. Similar to the first step, for each context clause, we will have three manual labels from the three annotators, and we adopt majority voting to determine its final label. The agreement rates for the types of context clauses are, on average, \underline{\textbf{88.91\%}}.

\subsubsection{Document Type}
\label{sec:documenttype}
Based on the above two steps, we have a label $y_c$ for each document indicating whether the contained ECP is conditional or not, and also the specific type labels for each context clause.


With $y_c$ and the pre-processed binary labels of all context clauses, we further define three types of documents for easier reference in the following sections. Specifically, a \textbf{Not-causal} type document contains a conditional ECP, but the document has no causal relationship since the context clauses needed by the conditional ECP are not included in the current document (i.e., $y_c = 1$); a \textbf{Conditional} type document contains a conditional ECP and its corresponding \textit{PR} context clauses indicating the condition (i.e., $y_c = 2$); and an \textbf{Others} type document contains a non-conditional ECP (i.e., $y_c = 0$).



After inspecting the dataset, we have $N_{Nc} = 146$ documents of Not-causal type, $N_{Con} = 763$ documents of Conditional type, and $N_O = 1176$ documents of Others type. 



\subsection{Negative Sampling}
\label{sec:negativesampling}
After manual annotation, we obtained the labels of conditional pairs and their context clauses. However, in the current dataset, all documents except for the ``Not-causal'' type are supposed to have valid causal relationships. This is due to that the conditional ECPs are all given their needed context clauses and the non-conditional ones do not depend on any context clause. In other words, the current dataset has much more ``positive" instances (i.e., $763+1176$ documents with causal relationships) than ``negative" instances (i.e., $146$ documents without causal relationships), which is too imbalanced to train a good classification model. Therefore, to generate more meaningful ``negative'' samples instead of simply down-sampling or up-sampling, we follow the procedure of negative sampling \cite{mikolov2013} and conduct the sampling step for all documents except for those of ``Not-causal'' type.

\subsubsection{Negative Sample Definition}
Specifically, we define the following two types of ``negative" samples. Firstly, a \textbf{Context-type} negative sample of a document is generated by replacing \textit{a subset of its original context} by another subset of a randomly sampled context from the other documents, while the ECP is kept unchanged. Then, an \textbf{Emotion-type} negative sample of a document is generated by replacing \textit{its emotion clause} by a randomly sampled emotion clause (indicating a different emotion) from the other documents, while the other clauses are kept unchanged.  




For the context-type negative sample, since ECPs contained in the document can be conditional or non-conditional, we further define different replacement operations for these two cases. Specifically, for documents with conditional ECPs, we want to highlight the \textit{PR} context clauses, so we can: (1) replace the \textit{PR} clauses while keeping the other context clauses unchanged to generate a new document without causal relationship; or (2) replace the other context clauses while keeping the \textit{PR} clauses unchanged to generate a new document with a causal relationship. As for the documents with non-conditional ECPs, we directly replace the whole context by a randomly sampled one from the other documents to generate a new document with a causal relationship.

Different from the context-type negative samples which may have different labels of a causal relationship, all emotion-type negative samples will not have valid causal relationships, since the replaced emotion clause indicates a different emotion.

\subsubsection{Number of Negative Samples}
After we define the types of negative samples and their corresponding sampling process, another important problem is how many negative samples we should generate for each original document. To generate a balanced dataset for our proposed task, we need to first calculate the numbers of documents with and without causal relationships after sampling. 

Suppose that for each document, we generate $n$ context-type negative samples and $n$ emotion-type negative samples. Specifically, since our main focus in this paper is on the documents of Condition type, for these documents, we generate $2n$ context-type samples, where $(n - \lfloor n / 2 \rfloor)$ samples are generated by replacing the \textit{IR} context clauses, and the rest $(n + \lfloor n / 2 \rfloor)$ samples are generated by replacing the \textit{PR} context clauses. Then, the number of documents with and without causal relationships in the constructed dataset can be calculated as follows:
\begin{equation}
\label{equ:determinen}
\begin{split}
N_{V} &= N_{Con} + N_{O}, \\
N_{pos} &= N_{V} + N_{Con} * (n - \lfloor \frac{n}{2} \rfloor) + N_{O} * n, \\
N_{neg} &= N_{Nc} + N_{V} * n + N_{Con} * (n + \lfloor \frac{n}{2} \rfloor),
\end{split}
\end{equation}
where $N_V$ denotes the number of instances with a valid causal relationship, $N_{pos}$ denotes the number of documents with causal relationships, and $N_{neg}$ denotes the number of documents without causal relationships.

To achieve a balanced dataset, we need to make sure that $\frac{N_{pos}}{N_{neg}} \approx 1$ from which we get:
\begin{equation}
N_{V} - N_{Nc} \approx N_{Con} * n + 2 * N_{Con} * (\lfloor n/2 \rfloor).
\end{equation} 
By assuming $n$ to be an even number or an odd number, the possible choices are $n = 2$ and $n = 3$. A smaller (or larger) $n$ may create an imbalanced dataset with too many positive (or negative) samples and cause the models biased toward the positive (or negative) labels. 
When $n$ is set to 2, we have 5554 positive documents and 5415 negative documents; while $n$ is set to 3, we have 7743 and 7668 positive and negative documents, respectively.
To study whether or not the value of $n$ will affect our proposed modules, we conduct experiments and report the results in Section \ref{sec:valueofn}, which indicates that the proposed approach can improve the performance of our task in both cases. 



\section{Model description}
\label{sec:model}
In this section, we introduce our end-to-end multi-task framework for the proposed task, in which we newly propose two general and effective modules to handle the two goals of the task accordingly, including a context masking module and a prediction aggregation module. 

\begin{figure*}
    \centering
    \includegraphics[width=0.9\linewidth]{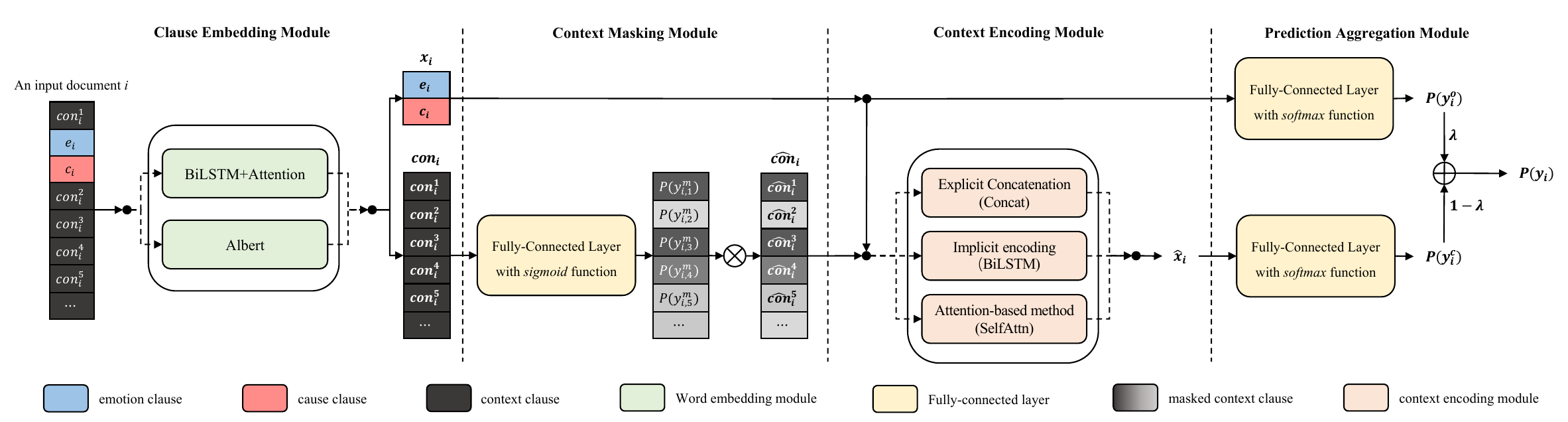}
    \caption{Architecture of our proposed framework, where the dash arrows denote the three possible choices for context encoding module, a darker entry of $P(y^m_i)$ denotes a larger value, and a lighter entry denotes a smaller value.}
    \label{fig:model}
\end{figure*}

\subsection{The Framework}
As shown in Figure \ref{fig:model}, our framework is mainly composed of four modules: a clause embedding module, a context masking module, a context encoding module, and a prediction aggregation module. In the clause embedding module, the word embedding vectors of the input clauses are passed into a Bi-directional Long Short-Term Memory (BiLSTM) model \cite{bilstm} to obtain clause embedding vectors for each clause. To extract the \textit{PR} context clauses for each document, in the context masking module, we create a mask over all context clauses to predict their probabilities of being \textit{PR} context clauses. After masking the context clauses, in the context encoding module, we encode context information into the clause embedding vectors of the input ECPs by using one of the three most classic methods: explicit concatenation, implicit encoding, and attention-based method. Finally, the context-encoded embedding vectors and the original ones of the input ECP are passed to the proposed prediction aggregation module to generate the final prediction. 

Below, we introduce the four modules in detail. Without the loss of generality, the formulas in subsequent discussions only consider the case where there is only one cause clause. Note that it can be easily extended to handle the cases with multiple cause clauses by concatenating their embedding vectors together. 

\subsection{Clause Embedding Module}
\label{sec:clause}
To obtain an embedding vector for each word, we propose two different schemes and compare their performance in the experiment. The first scheme is based on the word embedding vectors released by \cite{xia:19}, which are trained using the word2vec algorithm \cite{mikolov2013}, while the other one is constructed on a pre-trained language model to get each word's embedding vector. 

\subsubsection{Word Embedding (WE) by Word2Vec}
After getting each word embedding vector trained by word2vec algorithm \cite{xia:19}, we further adopt a word-level BiLSTM model and a fully-connected layer with an attention mechanism to encode words' embedding vectors into a clause embedding vector. The reason for choosing the BiLSTM model here is that it has been proven to be capable of generating an informative vector for each clause by passing words' information along the clause forwards and backward \cite{chencoling}.
Specifically, for an input clause $s=(v^1_s, \cdots, v^l_s)$ where $v^j_s$ denote the $j$-th word and $l$ denotes the maximum clause length, its embedding vector $\mathbf{s}$ is achieved following with a trainable weight matrix $W_{cls}$:
\begin{equation}
\begin{split}
&v^1_s, \cdots, v^l_s = BiLSTM(v^1_s, \cdots, v^l_s), \\ 
&\mathbf{s} = \sum^l_j \alpha_j v^j_s, \quad \alpha_j = \frac{\exp(W_{cls}v^j_s)}{\sum_k \exp(W_{cls}v^j_s)},
\end{split}
\end{equation}
where $\alpha_j$ denotes the attention value computed for the $j$-th word when achieving the clause embedding.

\subsubsection{Pre-trained Albert Model}
As for adopting a pre-trained language model, here we choose Albert \cite{lan2020albert}. The reason for adopting Albert instead of Bert is that Albert requires less memory space but achieves competitive performance compared to the Bert model, which can be easily adopted due to the lack of a GPU device with large memory. To get the clause embedding vector, specifically, we pass each clause separately to a loaded pre-trained Bert model to get the embedding vectors of each word and treat the embedding vector of the first word as the clause embedding vector. Also, considering the difference between our dataset and the data used for pre-training the Bert model, we add another layer of the transformer to the output of the loaded Bert model, which aims to learn task-specific semantic information to improve task performance. An extra fully-connected layer is added to align the dimension of the output vector from Albert and the dimension needed for the subsequent context encoding module. The clause embedding vector $\mathbf{s}$ for an input clause $s$ is achieved as shown below.
\begin{equation}
\begin{split}
v^1_s, \cdots, v^l_s &= Transformer(Albert(v^1_s, \cdots, v^l_s)), \\ 
\mathbf{s} &= FullyConnected(v^1_s)
\end{split}
\end{equation}

\subsubsection{Embedding vectors for all clauses}

For the $i$-th document, the input of this module includes three parts: the cause clause $c_i$, the emotion clause $e_i$, and the set of context clauses $con_i$. Following the above two schemes, we get $\mathbf{x}_i = [\mathbf{c}_i$;$\mathbf{e}_i$], and $\mathbf{con}_i=[\mathbf{con}^1_i, \cdots, \mathbf{con}^{L}_i]$ with $L$ being the maximum number of context clauses.

\subsection{Context Masking Module (CMM)}
\label{sec:contextmasking}
With the embedding vector of each clause, we create a trainable mask over all the context clauses for the purpose of learning the contributions of these context clauses to the causal relationship. Specifically, as shown by the following equation, for each context clause $con^j_i$, we pass it to a fully-connected layer with a sigmoid activation function to calculate its probability of being a \textit{PR} context clause $P(y^m_{i,j})$. Then, we multiply each $\mathbf{con}^j_i$ with its $P(y^m_{i,j})$ to get a weighted embedding vector $\mathbf{\widehat{con}}^j_i$, which takes into consideration how likely a context clause is essential for the causal relationship.
\begin{equation}
\begin{split}
&P(y^m_{i,j}) = sigmoid(W_{con}\cdot \mathbf{con}^j_i+b_{con}), \\
&\mathbf{\widehat{con}}^j_i = P(y^m_{i,j})\cdot \mathbf{con}^j_i, \quad 
\mathbf{\widehat{con}}_i = [\mathbf{\widehat{con}}^1_i, \cdots, \mathbf{\widehat{con}}^{L}_i],
\end{split}
\end{equation}
here, $W_{con}$ and $b_{con}$ are trainable weight matrix and bias terms of the fully-connected layer, respectively.

\subsection{Context Encoding Module}
\label{sec:contextencoding}
To determine the causal relationship of the input ECP under a specific context, we need to encode context information into the embedding vectors of the input ECP for the subsequent prediction. To this end, we consider three classic methods used most frequently in the area of text processing: explicit concatenation, implicit encoding, and attention-based method. The performance of these methods with our proposed modules will be compared and discussed in Section \ref{sec:experiment}.

\subsubsection{Explicit Concatenation}
This method directly concatenates the embedding vectors of the context clauses to those of the input pair and passes them to the next module so that the final prediction is based on all clauses, i.e., $\hat{\textbf{x}}_i = [\mathbf{c}_i;\mathbf{e}_i;\mathbf{\widehat{con}}_i]$.

\subsubsection{Implicit Encoding}
The second method encodes context information implicitly into the embedding vectors of the input ECP via an extra layer, such as BiLSTM or Convolutional Neural Network (CNN). Specifically, BiLSTM is capable of simulating the information propagation across a sequence of sentences \cite{chenemnlp}, while CNN has been proven to capture the latent pattern across the different local neighborhoods of sequence data features \cite{markcbm,BASIRI2021279,ma15010354}. Considering that the relevant information may be located anywhere in the context clauses, CNN may not be a good choice due to its fixed neighborhood size. Therefore, we adopt a clause-level BiLSTM model here for implicitly encoding context information and only focus on the resultant vectors of the cause and emotion clauses, i.e.,
\begin{equation}
\hat{\mathbf{c}}_i, \hat{\mathbf{e}}_i = BiLSTM(\mathbf{c}_i,\mathbf{e}_i,\mathbf{\widehat{con}}_i), \quad 
\hat{\textbf{x}}_i = [\hat{\mathbf{c}}_i;\hat{\mathbf{e}}_i].
\end{equation}

\subsubsection{Attention-based Method}
The third method is based on the self-attention module proposed by \cite{transformer}. We adopt the 1-layer-multi-head self-attention module to encode context information. Specifically, instead of calculating the attention scores among all sentences as in the original self-attention module, we only calculate the attention scores between the input pair and the context clauses. In this way, it reduces unnecessary attention weights and targets generating the context-encoded embedding vectors of the input ECP for the subsequent prediction. Specifically, we have:
\begin{equation}
\hat{\mathbf{c}}_i = \mathbf{c}_i + \sum_{j \in L}\alpha_{c, j} \cdot \mathbf{\widehat{con}}^j_i, \quad  
\hat{\mathbf{e}}_i = \mathbf{e}_i + \sum_{j \in L}\alpha_{e, j} \cdot \mathbf{\widehat{con}}^j_i,
\end{equation}
and $\hat{\textbf{x}}_i = [\hat{\mathbf{c}}_i;\hat{\mathbf{e}}_i]$, where,
\begin{equation}
\alpha_{u, j} = \frac{\exp(\mathbf{u}_i \cdot \mathbf{\widehat{con}}^j_i)}{\sum_{j' \in L}{\exp(\mathbf{u}_i \cdot \mathbf{\widehat{con}}^{j'}_i)}}, \quad u = \{c,e\},
\end{equation}
where $\alpha_{u,j}$ denotes the attention value between clause $u$ and clause $j$, and $u$ is a symbol variable whose value can be $c$ or $e$ with $c$ denoting cause and $e$ denoting emotion.

\subsection{Prediction Aggregation Module (PAM)}
\label{sec:pam}
As defined in Section \ref{sec:problem}, the conditional pairs will no longer have causal relationships if an irrelevant context is given, whereas the non-conditional pairs will always have valid causal relationships. Taking such a difference into consideration, we propose here a simple, general, and effective prediction aggregation module.

First, to get the prediction with context $y^c_i$, we pass the context-encoded embedding vectors of the input pair, $\hat{\mathbf{x}}_i$, to a fully-connected layer with a softmax activation function, $P(y^c_i) = softmax(W_c\hat{\mathbf{x}}_i))$, where $W_c$ is a trainable weight matrix. 

Next, we add an extra step of predicting the labels of causal relationships directly based on the original embedding vectors of the input pair, $y^o_i$, without encoding the context information. Specifically, we pass the original embedding vectors achieved in the clause embedding module, $\mathbf{x}_i$, to a fully-connected layer with a softmax activation function, $P(y^o_i) = softmax(W_o\mathbf{x}_i)$
, where $W_o$ is a trainable weight matrix. 

The proposed module works as follows. If $P(y^o_i)$ has already shown that the input pair has a valid causal relationship (i.e., $P(y^o_i = 1) > P(y^o_i = 0)$), then this pair is more likely to still have a causal relationship under any specific context, and the final result should depend more on the prediction without encoding context information. On the other hand, if the input pair is predicted to have no causal relationship without context, the final result should give more weight to the prediction that takes context information into consideration. Following this logic, we can have the aggregation formula below:
\begin{equation}
\label{eqa:agg}
\begin{split}
P(y_i) &= \lambda * P(y^o_i) + (1-\lambda) * P(y^c_i), \\
\lambda &= P(y^o_i = 1).
\end{split}
\end{equation}
This aggregation module allows the model to handle both conditional and non-conditional pairs, and gives a better prediction on the causal relationship of an input pair under a specific context.

\subsection{Model Training}
\label{sec:modeltraining}
As defined in Section \ref{sec:problem}, given an input document, our model has two objectives. Therefore, we propose a multi-task loss function for the model to simultaneously attend to these two objectives. 

\subsubsection{The Loss of Causal Relationship Classification Task}
\label{sec:classificationloss}
To supervise the learning of $P(y_i)$, we adopt the cross-entropy loss. Moreover, since the aggregation formula of $P(y_i)$ (Eq. (\ref{eqa:agg})) heavily depends on the prediction without context $P(y^o_i)$, we add another cross-entropy loss term to supervise this prediction. Therefore, we add these two loss terms up to jointly supervise the learning of the predictions with and without context. The loss function of the causal relationship classification tasks is:
\begin{equation}
\label{equ:lossp}
\begin{split}
    \mathcal{L}_{p} &= \mathcal{L}_{P(y^o_i)} + \mathcal{L}_{P(y_i)}, \\
    \mathcal{L}_{P(y)} &= -\frac{1}{D}\sum^D_{i=0}y\cdot \log(P(y)),
\end{split}
\end{equation}
where $D$ is the total number of documents in the dataset, $y$ is the ground-truth label, and $P(y)$ is the predicted probability. Note that the ground-truth value of $y^o_i$ can be transformed from $y_c$ achieved in Section \ref{sec:condlabel}, where $y^o_i = 1$ when $y_c = 0$ and $y^o_i = 0$ when $y_c > 0$. 

The reason of not adding one more cross-entropy loss in Eq. (\ref{equ:lossp}) to supervise $P(y^c_i)$ is that $P(y_i)$ and $P(y^c_i)$ share the same ground-truth labels, and adding another loss to supervise $P(y^c_i)$ may cause the overfitting problem. To validate such a setting, we conduct preliminary experiments and report the results with different combinations of $\mathcal{L}_{P(y)}$ in Section \ref{sec:validateL}, which indicates that adding $\mathcal{L}_{P(y^c_i)}$ leads to a decrease of the classification performance.

\subsubsection{The Loss of the Binary Mask Prediction Task} 
We adopt an element-wise binary cross-entropy loss to supervise the learning of the mask over all the context clauses. Note that there are many more documents without \textit{PR} context clauses than those with \textit{PR} context clauses in the constructed dataset (i.e., around 85\% of the masks have value 0 in all positions and only 15\% of the masks have value 1 in some positions). If we calculate the loss of all masks, the predictions will be easily biased to 0. Therefore, we only calculate the loss of those documents with \textit{PR} context clauses. Specifically, we have:
\begin{equation}
\begin{split}
    \mathcal{L}_{m} &= -\frac{1}{D_{PR}}\sum^{D_{PR}}_{i=0}\frac{1}{L}\sum_{j\in L}\mathcal{L}_{m_{i,j}}, \\
    \mathcal{L}_{m_{i,j}} &= y^m_{i,j}\cdot \log(P(y^m_{i,j})) + (1-y^m_{i,j})\cdot \log(1-P(y^m_{i,j})),
\end{split}
\end{equation}
where $D_{PR}$ denotes the number of documents with \textit{PR} context clauses, $y^m_{i,j}$ is the ground-truth binary label indicating whether this context clause is a \textit{PR} context clause, and $P(y^m_{i,j})$ is the predicted probability achieved in Section \ref{sec:contextmasking}.

\subsubsection{The Multi-task Loss Function}
\label{sec:multitaskloss}
Based on the loss functions of the two tasks described above, the final loss function for our proposed model is:
\begin{equation}
\label{equ:loss}
\begin{split}
    \mathcal{J}(\theta) = \eta \mathcal{L}_{p} + \tau \mathcal{L}_{m} + \gamma\|\theta\|^2,
\end{split}
\end{equation}
where $\eta$ and $\tau$ are the loss weights of causal relationship recognition and mask prediction, respectively, and $\theta$ denotes all trainable parameters. Note that the final term is the regularization term and $\gamma$ is the regularization weight. To find the best values for $\eta$ and $\tau$, we conduct a grid search over different combinations of these two hyperparameters and report the results in Section \ref{sec:valueofetaandtau}. 


\section{Experiment}
\label{sec:experiment}
In Section \ref{sec:dataset}, we have described the process of dataset construction and the details of the constructed dataset\footnote{The constructed dataset and our programs can be found in: 
\url{https://drive.google.com/file/d/1thf5pPT73pMYgm2AeWq5c4sFzEdc-J76/view?usp=share\_link}}. In this section, we conduct experimental studies to evaluate our approach and analyze the experiment results pragmatically.

\subsection{Baseline Models}
As mentioned in Section \ref{sec:contextencoding}, there are three options in the context encoding module. Therefore, we consider three baseline models without the two proposed modules, each of which contains one of the three context encoding methods we have described. 
 \begin{itemize}
\item \textbf{Word Embedding based BiLSTM or Albert + Concatenation (WE-CC or AB-CC)}: this baseline model uses BiLSTM with word2vec word embeddings, or uses Albert at the word level, then directly concatenates the context clauses' vectors to those of the input pair.
\item \textbf{Word Embedding based BiLSTM or Albert + BiLSTM (WE-BL or AB-BL)}: this model uses BiLSTM with word2vec word embeddings or Albert at word level and uses BiLSTM at clause level to get the context-encoded embedding vectors of the input pair for the final prediction.
\item \textbf{Word Embedding based BiLSTM or Albert + Self-Attention (WE-SA or AB-SA)}: this model uses BiLSTM with word2vec word embeddings or Albert at word level and uses Self-Attention at clause level to encode the context information.
\end{itemize}

\subsection{Experiment Settings}
\label{sec:experimentsetting}
To avoid the effect of randomness, we divide the whole dataset into 5 folds and repeat the experiments 5 times with each fold being the testing data. The average experiment results are reported in the following sections. 

To evaluate the task performance, we adopt different metrics for the two goals in our proposed task. Specifically, for binary causal relationship classification, we use the traditional precision (\textbf{P}), recall (\textbf{R}), and \textbf{F1} scores. 

As for the task of finding the \textit{PR} context clauses, the classification threshold of the sigmoid function is set to $\mathbf{0.1}$ due to the extremely imbalanced labels of \textit{PR} context clauses, where $\mathbf{0.1}$ is determined according to the cross-validation results. To evaluate the performance, we define the following metrics: 
\begin{itemize}
    \item Global F1 value (\textbf{gF1}): We calculate the F1 value for the predictions of all \textit{PR} context clauses contained in the test data;
    \item Document-level F1 value (\textbf{dF1}): This metric differs from the global one in that it first calculates the F1 value of the \textit{PR} context clauses' predictions for each test document containing these context clauses, and then averages them;
    \item Ratio of All-correct Cases (\textbf{rAC}): We calculate the ratio of the test samples whose predictions are all correct out of the test samples that have \textit{PR} context clauses;
    \item Overall Accuracy (\textbf{Acc}): We calculate the overall average accuracy of the mask prediction for all test samples, including both documents with and without \textit{PR} context clauses.
\end{itemize}
Note that \textbf{gF1}, \textbf{dF1}, and \textbf{rAC} are calculated only for the test samples with \textit{PR} context clauses, while \textbf{Acc} is calculated based on all test samples.

Besides the evaluation metrics, we also provide some detailed settings of our model here. The adopted pre-trained Albert model is published by Google research team\footnote{https://github.com/google-research/albert}. The hidden units in BiLSTM is set to 100, and the number of heads in the self-attention module is set to 1. All weight matrices are randomly initialized with uniform distribution. For training of WE-based baselines, we use the stochastic gradient descent algorithm and Adam optimizer, with batch size set to 128 and learning rate set to 0.001. As for the training of the Albert-based baselines, we adopt the same optimizer used for pretraining the Albert model, the LAMBOptimizer \cite{you2020large}, and the learning rate is set to $10^{-4}$. Also, for regularization, dropout is applied with a dropout rate set to 0.2, and an L2-norm regularization term is added to constraint the trainable parameters, where the weight of the regularization term $\gamma$ is set to $10^{-5}$ empirically.

In what follows, we first report the preliminary experimental results to validate some specific settings adopted in model design and the main experiments, and then analyze the performance of the models with and without our proposed modules.

\begin{table}
\centering 
\caption{Task performance under different $n$}
\label{tab:validaten}
\renewcommand\arraystretch{1.4}
\resizebox{75mm}{34mm}{
\begin{threeparttable}
\begin{tabular}{clcc}
\toprule
Setting & \multicolumn{1}{c}{Model} & \textbf{F1} & \textbf{gF1} \\
\midrule
\multirow{6}{*}{$n$ = 2} & WE-CC & 0.6264 & - \\
& WE-BL & 0.6883 & - \\
& WE-SA & 0.6305 & - \\
& WE-CC+C+P & \textbf{0.7338} ($\uparrow$ 0.1074) & 0.7043\\
& WE-BL+C+P & \textbf{0.7624} ($\uparrow$ 0.0741) & 0.7021 \\
& WE-SA+C+P & \textbf{0.7716} ($\uparrow$ 0.1411) & 0.7079\\
\midrule
\multirow{6}{*}{$n$ = 3} & WE-CC & 0.6728 & - \\
& WE-BL & 0.7305 & - \\
& WE-SA & 0.7357 & - \\
& WE-CC+C+P & \textbf{0.7408} ($\uparrow$ 0.0680) & 0.7897 \\
& WE-BL+C+P & \textbf{0.7932} ($\uparrow$ 0.0627) & 0.7917\\
& WE-SA+C+P & \textbf{0.7985} ($\uparrow$ 0.0628) & 0.795 \\
\bottomrule
\end{tabular}
\begin{tablenotes}
\footnotesize
\item[.] ``+C'' and ``+P'' stand for the addition of CMM and PAM modules, respectively.
\item[.] ``-'' denotes no performance value.
\end{tablenotes}
\end{threeparttable}
}
\end{table}

\subsection{Preliminary Results}
We report in this section the preliminary experimental results to validate the settings of the value of $n$ (Section \ref{sec:negativesampling}), the loss function of the classification task $\mathcal{L}_p$ (Section \ref{sec:classificationloss}), and the values of $\eta$ and $\tau$ (Section \ref{sec:multitaskloss}). For simplicity, we only conduct the preliminary experiments on WE-based baselines, as the best parameter setting of WE-based baselines should also be applicable to Albert-based baselines.

\begin{figure*}
    \centering
    \includegraphics[width=0.8\textwidth]{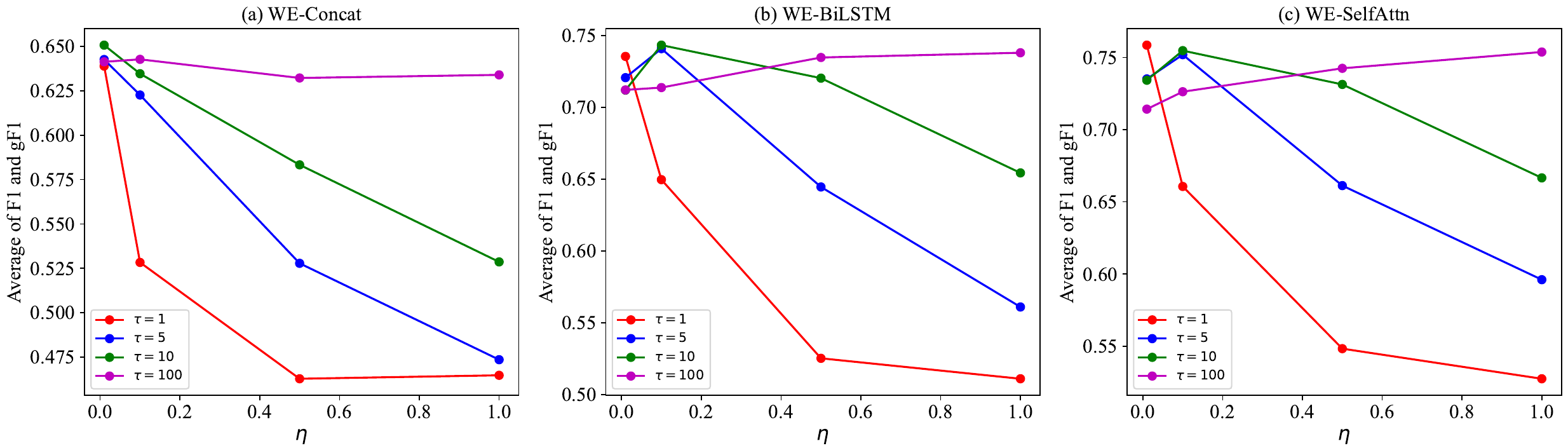}
    \caption{Average Performance of adding CMM to the three WE-base models under different $\eta$ and $\tau$}
    \label{fig:losssearch}
\end{figure*}

\subsubsection{Effect of $n$}
\label{sec:valueofn}
To study the influence of $n$, we conduct experiments on the constructed dataset by setting $n$ to 2 and 3 and report the experiment results in Table \ref{tab:validaten}. We notice that in both cases, the models with our proposed modules can consistently achieve higher \textbf{F1} values than the baselines without the proposed modules in the causal relationship classification task. In terms of the performance of finding the \textit{PR} context clauses, all models can achieve good performances (i.e., high \textbf{gF1} values). These observations indicate that the value of $n$ will not affect the generality and effectiveness of our proposed modules. For simplicity, we utilize the dataset with $n=2$ for our comparative experiments and ablation studies.


\begin{table}
\centering 
\caption{Classification performance under different combination of $\mathcal{L}$}
\label{tab:validateL}
\renewcommand\arraystretch{1.4}
\resizebox{78mm}{42mm}{
\begin{threeparttable}
\begin{tabular}{clc}
\toprule
Model & \multicolumn{1}{c}{Loss Terms} & \textbf{F1} \\
\midrule
WE-CC & $\mathcal{L}_{P(y_i)}$ & 0.6264 \\ 
\midrule
\multirow{4}{*}{WE-CC+P} & $\mathcal{L}_{P(y_i)}$ & 0.6439 ($\uparrow$ 0.0175) \\
& $\mathcal{L}_{P(y_i)} + \mathcal{L}_{P(y^o_i)}$ & \textbf{0.7138 ($\uparrow$ 0.0874)} \\
& $\mathcal{L}_{P(y_i)} + \mathcal{L}_{P(y^c_i)}$ & 0.6237 ($\downarrow$ 0.0027) \\
& $\mathcal{L}_{P(y_i)} + \mathcal{L}_{P(y^o_i)} + \mathcal{L}_{P(y^c_i)}$ & 0.6995 ($\uparrow$ 0.0731) \\
\midrule
\midrule
WE-BL & $\mathcal{L}_{P(y_i)}$ & 0.6883 \\
\midrule
\multirow{4}{*}{WE-BL+P} & $\mathcal{L}_{P(y_i)}$ & 0.6605 ($\downarrow$ 0.0278) \\
& $\mathcal{L}_{P(y_i)} + \mathcal{L}_{P(y^o_i)}$ & \textbf{0.7194 ($\uparrow$ 0.0311)}  \\
& $\mathcal{L}_{P(y_i)} + \mathcal{L}_{P(y^c_i)}$ & 0.6582 ($\downarrow$ 0.0301) \\
& $\mathcal{L}_{P(y_i)} + \mathcal{L}_{P(y^o_i)} + \mathcal{L}_{P(y^c_i)}$ & 0.7064 ($\uparrow$ 0.0181) \\
\midrule
\midrule
WE-SA & $\mathcal{L}_{P(y_i)}$ & 0.6305 \\
\midrule
\multirow{4}{*}{WE-SA+P} & $\mathcal{L}_{P(y_i)}$ & 0.6563 ($\uparrow$ 0.0258) \\
& $\mathcal{L}_{P(y_i)} + \mathcal{L}_{P(y^o_i)}$ & \textbf{0.7009 ($\uparrow$ 0.0704)} \\
& $\mathcal{L}_{P(y_i)} + \mathcal{L}_{P(y^c_i)}$ & 0.6527 ($\uparrow$ 0.0222) \\
& $\mathcal{L}_{P(y_i)} + \mathcal{L}_{P(y^o_i)} + \mathcal{L}_{P(y^c_i)}$ & 0.6971 ($\uparrow$ 0.0666) \\
\bottomrule
\end{tabular}
\begin{tablenotes}
\item[.] ``+P'' denotes the addition of the PAM module.
\item[.] The bold values are the best performances.
\item[.] $\uparrow$ and $\downarrow$ denote the increase and decrease of the performance compared with the baseline without PAM.
\end{tablenotes}
\end{threeparttable}
}
\end{table}

\subsubsection{Effect of Different Loss Terms in $\mathcal{L}_p$}
\label{sec:validateL}

As described in Section \ref{sec:modeltraining}, when calculating the causal relationship classification loss $\mathcal{L}_p$, we only consider the sum of $\mathcal{L}_{P(y_i)}$ and $\mathcal{L}_{P(y^o_i)}$, leaving out $\mathcal{L}_{P(y^c_i)}$. To validate such a setting, we conduct experiments on the models with PAM and train them using different combinations of the loss terms. The results are shown in Table \ref{tab:validateL}. 

As we can easily observe from the table, if we only add PAM to the base models without adding corresponding loss term (i.e., only using $\mathcal{L}_{P(y_i)}$ for ``Baselines+PAM''), the performance is not stable. On the other hand, if we train the models with all loss terms, the performance is improved compared with the baselines without PAM. More interestingly, when we remove $\mathcal{L}_{P(y^c_i)}$, we observe that the performance is even better than that of training with all loss terms, while the performance is dramatically decreased if we remove $\mathcal{L}_{P(y^o_i)}$. 

Such observations demonstrate that the supervision of the prediction without context information (i.e., adding $\mathcal{L}_{P(y^o_i)}$) is a necessary part if we want to utilize PAM, and the supervision towards the prediction with context information (i.e., adding $\mathcal{L}_{P(y^c_i)}$) may be redundant and not necessary. Therefore, we do not introduce the cross-entropy loss term $\mathcal{L}_{P(y^c_i)}$ in the loss function of the causal relationship classification task.

\subsubsection{Effect of $\eta$ and $\tau$}
\label{sec:valueofetaandtau}

During training, we observe that the loss of the mask prediction task is much smaller than that of the causal relationship classification task (i.e., $\mathcal{L}_p \gg \mathcal{L}_m$). This is due to that in around 85\% of the masks the values are 0, and we only calculate $\mathcal{L}_m$ on those masks with values 1. 
Based on this observation, we need to set a small value for $\eta$ and a large value for $\tau$ in Eq. (\ref{equ:loss}). Specifically, we have $\eta \in \{0.01, 0.1, 0.5, 1\}$ and $\tau \in \{1, 5, 10, 100\}$.

Figure \ref{fig:losssearch} displays the curves of the average performance values (i.e., (\textbf{F1} + \textbf{gF1}) / 2) under different combinations of $\eta$ and $\tau$ for different WE-base models. As shown by the figure, in general, a larger $\tau$ together with a smaller $\eta$ can yield good performance in both tasks, while the performance will be significantly decreased when $\tau = \eta$. 

Among all the combinations, setting $\eta$ to $0.1$ and $\tau$ to $10$ can generally achieve the best performance in both tasks for all the models, except for the WE-Concat model, which achieves the best performance under $\eta=0.01$ and $\tau=10$. Therefore, we set $\eta$ to $0.1$ and $\tau$ to $10$ for the Albert-based baselines in our following comparative experiments and ablation studies.

\subsection{Experiment results}

\subsubsection{Effect of PAM}
\label{sec:effectofpam}
We first focus on the effect of PAM, which only concerns the performance of causal relationship classification. As shown in Table \ref{tab:cmmpam}, compared with the three WE-based baselines without PAM, those with PAM consistently achieve higher \textbf{F1} values, and the improvement is around 6\% on average. Specifically, WE-BiLSTM achieves the best performance, and the improvement of the Concat, BiLSTM, and SelfAttn models is around 8\%, 3\%, and 6\%, respectively. Such a difference is possibly due to that the self-attention module in SelfAttn may need a better feature extraction module at the word level to train well, while simple concatenation utilized by Concat cannot get semantic embeddings encoded with context information. 

As for the Albert-based baselines, we observe that the highest performance is achieved by the Albert-Concat model, followed by Albert-SelfAttn and finally Albert-BiLSTM. Interestingly, when equipped with a stronger language model, BiLSTM performs slightly worse in encoding context information, which can be ascribed to the conflict between the strong pre-trained semantic knowledge brought by Albert and the task-specific context knowledge learned by BiLSTM. 

The above results and analysis demonstrate that PAM is general enough to be used together with existing classic models and it can improve the performance of the proposed causal relationship classification task, whose improvement depends on how well the context information is encoded.

\begin{table}
\centering 
\caption{Task performance comparison}
\label{tab:cmmpam}
\renewcommand\arraystretch{1.5}
\resizebox{\linewidth}{55mm}{
\begin{threeparttable}
\begin{tabular}{lcccccccc}
\toprule
\multirow{2}{*}{Model} & \multicolumn{3}{c}{\tabincell{c}{Causal Relationship \\Classification}} & \multicolumn{4}{c}{Mask Prediction} \\
\cmidrule(lr){2-4} \cmidrule(lr){5-8}
& \textbf{P} & \textbf{R} & \textbf{F1} & \textbf{gF1} & \textbf{dF1} & \textbf{rAC} & \textbf{Acc} \\
\midrule
WE-CC & 0.5746 & 0.6936 & 0.6264 & - & - & -  & - \\
WE-CC+P & 0.6647 & 0.7851 & 0.7138 & - & - & - & - \\
WE-CC+C & 0.5910 & 0.6557 & 0.6206 & 0.6809 & 0.6378 & 91/305  & 0.8811 \\
WE-CC+C+P & 0.6702 & 0.8108 & 0.7338 & 0.7043 & 0.6555 & 93/305  & 0.8812 \\
AB-CC & 0.5563 & 0.8220 & 0.6630 & - & - & - & - \\
AB-CC+P & 0.6701 & 0.8047 & 0.7312 & - & - & - & - \\
AB-CC+C & 0.6020 & 0.5642 & 0.5823 & 0.6723 & 0.6398 & 85/305 & 0.8784 \\
AB-CC+C+P & \textbf{0.6905} & \textbf{0.8351} & \textbf{0.7557} & \textbf{0.7198} & \textbf{0.6824} & \textbf{114/305} & \textbf{0.9006} \\
\midrule
WE-BL & 0.6736 & 0.7041 & 0.6883 & - & - & - & - \\
WE-BL+P & 0.6594 & 0.7918 & 0.7194 & - & - & - & - \\
WE-BL+C & \textbf{0.7652} & 0.8057 & \textbf{0.7846} & 0.7021 & 0.6583 & 97/305 & 0.8877 \\
WE-BL+C+P & 0.6873 & \textbf{0.8560} & 0.7624 & 0.7021 & 0.6584 & 96/305 & 0.8879 \\
AB-BL & 0.5749 & 0.8037 & 0.6694 & - & - & - & - \\
AB-BL+P & 0.6624 & 0.7864 & 0.7190 & - & - & - & - \\
AB-BL+C & 0.6730 & 0.6751 & 0.6741 & \textbf{0.7298} & \textbf{0.6863} & \textbf{118/305} & 0.8987 \\
AB-BL+C+P & 0.6889 & 0.8320 & 0.7536 & 0.7279 & 0.6810 & 113/305 & \textbf{0.9048} \\
\midrule
WE-SA & 0.5905 & 0.6847 & 0.6305 & - & - & - & - \\
WE-SA+P & 0.6793 & 0.7177 & 0.6971 & - & - & - & - \\
WE-SA+C & \textbf{0.7643} & 0.8362 & \textbf{0.7985} & 0.7185 & 0.6648 & 102/305 & 0.8874 \\
WE-SA+C+P & 0.6899 & \textbf{0.8756} & 0.7716 & 0.7079 & 0.6684 & 106/305 & 0.8853\\
AB-SA & 0.5292 & 0.8536 & 0.6529 & - & - & - & - \\
AB-SA+P & 0.6897 & 0.7643 & 0.7296 & - & - & - & - \\
AB-SA+C & 0.7511 & 0.7784 & 0.7644 & 0.7216 & 0.6670 & 114/305 & 0.9014 \\
AB-SA+C+P & 0.6827 & 0.8540 & 0.7587 & \textbf{0.7367} & \textbf{0.6787} & \textbf{133/305} & \textbf{0.9031}\\
\bottomrule
\end{tabular}
\begin{tablenotes}
\footnotesize
\item[.] ``+C'' and ``+P'' stand for the addition of CMM and PAM modules, respectively.
\item[.] ``-'' denotes no performance value.
\item[.] 305 is the average number of documents with \textit{PR} context clauses in the testing set.
\end{tablenotes}
\end{threeparttable}
}
\end{table}

\subsubsection{Effect of CMM}
Next, we focus on the effect of CMM. The results of adding the CMM to the three baselines are also reported in Table \ref{tab:cmmpam}, and we analyze the performance of the models from two perspectives: causal relationship classification and mask prediction. 

Firstly, we discuss the performance of causal relationship classification. From the table, we can see that using CMM together with PAM can guarantee an improvement on the \textbf{F1} score compared with only using PAM while using CMM only is shown to have different effects on different base models. Specifically, for the WE-BiLSTM, WE-SelfAttn, and Albert-SelfAttn models, only adding CMM achieves the best classification performance, which is even better than that of using CMM with PAM. We think this is due to the similar mechanism of CMM and PAM, which will be elaborated further in Section \ref{sec:discusscmmpam}. 

As for the rest baseline models, only adding CMM may not improve the classification performance. The reason behind the two Concat models may be that the Concat model cannot encode context information well, while the Albert-BiLSTM model may be a trade-off between the losses of the two tasks. As analyzed in Section \ref{sec:valueofetaandtau}, we choose $\eta = 0.1$ and $\tau = 10$ for all models according to their average performance. However, it is possible that a parameter search can also help Albert-BiLSTM to find its best performance when it works with the proposed CMM module. As our main focus is not to tune and find the best results for Albert-based models, we leave this exploration for future work. From the above demonstrations, we conclude that in general, CMM can promote the effect of encoding context information and benefit the recognition of causal relationships.

Secondly, we discuss the performance of the mask prediction task (i.e., finding the \textit{PR} context clauses). We calculate the 4 metrics of mask prediction as defined in Section \ref{sec:experimentsetting}. Note that in each fold of the dataset, we have on average 305 documents with \textit{PR} context clauses. As shown in the table, CMM can generally achieve high \textbf{gF1} values and \textbf{dF1} values when combined with all three baselines. 
The average \textbf{rAC} is only around 105 out of 305, which is naturally understandable due to the difficulty of the task of correctly predicting context clauses.
Moreover, to avoid the situation that CMM directly outputs value 1 in all positions during prediction, we further calculate the overall accuracy \textbf{Acc} for all test samples. We can see that the average \textbf{Acc} is pretty high ($0.8914$). Therefore, we conclude that the CMM module can provide good performance in mask prediction, and it can be easily combined into most existing models as an auxiliary module to find the specific context clauses participating in the causal relationships.

\subsection{Discussion of CMM and PAM}
\label{sec:discusscmmpam}
As aforementioned, we observe that using CMM only may achieve a better performance than using CMM together with PAM in the causal relationship classification task. CMM can filter out irrelevant context clauses, and help the models to focus on learning the dependencies between the ECP and those remaining context clauses. It already enables the model to dynamically change its focus during prediction according to whether or not the ECP is conditional, like what PAM does. Therefore, for baseline models with a relatively stronger ability to encode context information like BiLSTM and SelfAttn, using CMM only can get the best performance due to the removal of the noisy context clauses, while using CMM with PAM may cause an overfitting problem. As for the Concat model, the improvement brought by adding CMM is less significant and less stable, because the Concat model does not consider the dependencies among clauses during prediction, and the filter effect of CMM may instead reduce the information contained in the input of the model. Therefore, for the Concat model, using CMM with PAM is the best choice, where PAM plays its original role to fine-tune the prediction results and CMM focuses on improving the predictions with context information.

To sum up, if the baseline models adopt a context encoding module learning the dependencies among clauses during prediction, adding CMM only can achieve good performance for our proposed task. If the adopted context encoding module is relatively simple, using CMM with PAM is a better choice.

\subsection{Ethical Consideration}
The dataset used in this work was originally proposed by \cite{xia:19}, who have already taken the necessary steps to avoid privacy violations. Specifically, we have further paraphrased all the examples mentioned in this paper following the moderate disguise scheme \cite{Bruckman2002}. We claim that we only focus on investigating the conditional causal relationships from text data and do not make any interference or diagnosis towards anyone’s experiences.

\section{Conclusion and Future Work}
\label{sec:conclusion}
In this paper, we articulate the importance of context in determining the causal relationships between emotions and their causes. To achieve such a goal, we define a new task with two steps. The first step is to determine whether or not an input emotion-cause pair has a causal relationship under a specific context, aiming to judge whether a causal condition is needed and whether it is contained in the context. If a causal condition is needed, the second step is to extract the context clauses that participate in the causal relationship as the targeted causal condition. We construct a dataset through manual annotation and negative sampling based on the ECPE dataset, containing the annotations of conditional causal relationships and the causal conditions. Furthermore, we propose a multi-task framework with a context masking module (CMM) and a prediction aggregation module (PAM) to achieve the two goals of our task. Experiments demonstrate that our proposed modules can significantly improve the performance of our proposed task, and enjoy high generality so that they can be easily combined into most existing models. This work is a pilot study viewing the importance of context in conditional causal relationships. There remain many important and interesting problems ahead of us. For example, quantifying the effect of context on the targeted causal relationship is an important task in studying this conditional causal relationship and modeling the contextual information within a document \cite{li-etal-2023-recurrent}. Besides, how to enable the existing emotion-cause pair extraction models to take the effect of context into account is another interesting task. In the context of large language models, we can explore the causal reasoning capacity of various large language models \cite{li2023label} on our constructed dataset.

\ifCLASSOPTIONcaptionsoff
  \newpage
\fi



%
\bibliographystyle{IEEEtran}
\bibliography{IEEEabrv,mytac}




\end{document}